\documentclass[mlabstract]{jmlr}

\jmlrproceedings{NeurReps}{NeurReps 2026 -- Extended Abstract Track}

\makeatletter
\renewcommand*{\@jmlrenddoc}{%
  \FloatBarrier
  \global\let\@reprint\@empty
}
\makeatother

\usepackage{bm}
\usepackage{enumitem}
\usepackage{cleveref}

\usepackage{longtable}%

\usepackage{booktabs}
\usepackage[load-configurations=version-1]{siunitx} %

\newcommand{\R}{\mathbb{R}}
\newcommand{\E}{\mathbf{E}}

\newcommand{\J}{\mathbf{J}}
\newcommand{\G}{\mathbf{G}}
\newcommand{\bx}{\boldsymbol{x}}
\newcommand{\bu}{\boldsymbol{u}}
\newcommand{\bz}{\boldsymbol{z}}

\renewcommand{\top}{\intercal}

\newcommand{\T}{\mathcal{T}} 

\theorembodyfont{\upshape}
\theoremheaderfont{\bfseries}
\theorempostheader{.}
\theoremsep{}

\jmlrvolume{}
\firstpageno{1}
\editors{}

\jmlryear{2026}
\jmlrworkshop{Symmetry and Geometry in Neural Representations}

\title[Beyond Local Linearity]{
Beyond Local Linearity: \\ 
Scale-Resolved Geometry of Learned Image Encoders
}

\author{\Name{Jakub Szymkowiak\nametag{$^{1,2}$}}\and
  \Name{Wojtek Pa{\l}ubicki\nametag{$^{1}$}}\and
  \Name{Kamil Adamczewski\nametag{$^{1,2}$}}\\
  \addr $^{1}$Adam Mickiewicz University in Pozna{\'n}, Poland\\
  \addr $^{2}$IDEAS NCBR, Warsaw, Poland}

\begin{document}

\maketitle

    \begin{abstract}
    Understanding how learned representations respond to finite input changes is important for characterizing their sensitivity, invariances, and robustness. Yet existing geometric analyses are predominantly local and describe only infinitesimal perturbations. We introduce a scale-resolved statistic that compares an encoder's measured feature displacement with its local linear prediction as the perturbation magnitude increases. Across diverse image encoders, we discover a characteristic plateau--rise--peak--decay profile, which we call the \emph{bump}. The bump is absent at initialization, emerges early during standard training, and does not form under randomized labels or random-noise inputs. Its shape also varies with the training distribution and robustness objective. These results establish departures from local geometry as a signature of how encoder representations are shaped by learning.

\end{abstract}
\begin{keywords}
pullback metric, neural representations, local geometry, linearization, vision
\end{keywords}

\section{Introduction}
\label{sec:01}

Deep neural networks are often studied through the geometry of their learned representations. During training, an encoder becomes selectively sensitive to input variation: it may suppress differences that are irrelevant to the objective while preserving those needed for prediction. In image classification, for example, variations within a class need not remain distinguishable if they do not affect the predicted label, whereas variations separating classes must be retained. This selective sensitivity shapes the geometry of the learned representation map.
Although there exist differential quantities describing the geometric imprint of this selectivity on the learned mapping \citep{novak2018sensitivity, arvanitidis2021latent}, how far beyond the immediate neighborhood of an input these descriptions predict feature displacement has not been measured.

In this work, we propose to measure how far this local geometric description remains predictive. To this end, we introduce a scale-dependent statistic that quantifies how the size of the encoder's actual feature displacement departs from its local linear prediction. Starting from an image, we gradually increase the perturbation magnitude and compare the size of the measured response with that predicted from the encoder's infinitesimal sensitivity, averaging over random perturbation directions. This reveals both the scale at which nonlinear behavior becomes important and the manner in which it emerges across architectures, training regimes, and evaluation data.

We find that all trained encoders considered in our study exhibit a characteristic scale-dependent profile: the measured feature displacement initially agrees in magnitude with its local linear prediction, then grows larger than predicted, reaches a maximum discrepancy at an intermediate scale, and eventually falls relative to the prediction. We refer to this structure as the \emph{bump}.

The bump is not produced by optimization alone. Networks trained with randomized labels, or on unstructured random inputs, do not develop a comparable peak, despite being optimized to fit their respective training objectives. These controls associate the bump with learning shared, meaningful input--target structure rather than with parameter changes or memorization alone.

\paragraph{Contributions}
\begin{itemize}[topsep=-0.03em, itemsep=-0.03em]
\item We introduce a scale-resolved approach for studying how the finite behavior of an image encoder departs from its local linear geometry.

\item We identify the \emph{bump}, a characteristic response profile that appears consistently across trained image encoders and emerges early during meaningful training.

\item 
Through controlled experiments, we show that the bump reflects how representations are shaped by the learning problem: it is absent under random-label memorization and unstructured data, and varies systematically with the training distribution and robustness objective.
\end{itemize}
\section{Method}
\label{sec:02}

Let $\Phi:\mathcal{I}\rightarrow\mathcal{Z}$ be a frozen image encoder mapping an image $\bx\in\mathcal{I}\subseteq[0,1]^{d_{\mathcal I}}$ to a feature representation $\Phi(\bx)\in\mathcal{Z}\simeq\mathbb{R}^{d_{\mathcal Z}}$.
We are interested in the encoder's sensitivity, that is, in how the representation changes as the input moves away from $\bx$.
To first order, this sensitivity is captured by the pullback metric $\G_\Phi(\bx) := \J_\Phi(\bx)^\top \J_\Phi(\bx)$, where $\J_\Phi(\bx)$ is the encoder's Jacobian at $\bx$.
Our goal is to determine how far this local description remains predictive as we move away from $\bx$.
We therefore compare the actual change in the representation under a finite perturbation with the change predicted by the metric.
By repeating this comparison across perturbation scales, we obtain a scale-resolved description of the encoder's departure from its local linear geometry.

For a unit direction $\bu\in\mathbb{S}^{d_{\mathcal I}-1}$ and a sufficiently small perturbation magnitude $\eta > 0$, the pullback metric predicts the squared feature displacement
    $
    D_\eta(\bx,\bu)
        :=
        \left\|
            \Phi(\bx+\eta\bu)-\Phi(\bx)
        \right\|_2^2
        \label{eq:finite_response}
    $
    of the perturbation $\bx \to \bx + \eta\bu$ as 
    $
        \eta^2\, \bu^\top \G_\Phi(\bx)\, \bu = \eta^2\left\|\J_\Phi(\bx)\bu\right\|_2^2.
        \label{eq:predicted_displacement}
    $   
    Equivalently, this can be seen by squaring both sides of the first-order step
    $
        \Phi(\bx+\eta\bu) - \Phi(\bx) \approx \eta\, \J_\Phi(\bx)\bu.
        \label{eq:first_order_approximation}
    $
    
This prediction, however, need not agree with the encoder's actual response $D_{\eta}(\bx, \bu)$. 
To reduce the comparison to a single direction-free measure, we average over isotropically sampled unit directions
$\bu\sim\operatorname{Unif}(\mathbb{S}^{d_{\mathcal I}-1})$
and define
\begin{equation}
    r_\eta(\bx)
    :=
    \frac{
        \mathbb{E}_{\bu}
        \left[
            D_\eta(\bx,\bu)
        \right]
    }{
        \eta^2\,
        \mathbb{E}_{\bu}
        \left[
            \left\|
                \J_\Phi(\bx)\bu
            \right\|_2^2
        \right]
    }.
    \label{eq:r_eta}
\end{equation}
The numerator of \cref{eq:r_eta} measures the encoder's mean response at scale $\eta$, whereas the denominator measures the response predicted by its local geometry at $\bx$. 
Thus, $r_\eta(\bx)\approx1$ indicates agreement in magnitude with the local linear prediction, while $r_\eta(\bx)>1$ and
$r_\eta(\bx)<1$ indicate a larger or smaller response than predicted, respectively. For an encoder differentiable at $\bx$ with nonzero local sensitivity $\J_{\Phi}(\bx) \neq \mathbf{0}$
$
    \lim_{\eta\rightarrow0}r_\eta(\bx)=1
$
by construction. 
The details on the finite-sample estimator of \cref{eq:r_eta} and the leading-order deviation from $r_{\eta} \approx 1$ are given in \appendixref{app:theory}.
\section{Experiments}
\label{sec:03}

We study whether scale-dependent departures from local geometry follow a
consistent pattern across image encoders, how they emerge during training, and which properties of the learning problem determine their formation.
Unless stated otherwise, we evaluate $r_\eta$ at $17$ logarithmically
spaced perturbation scales and report the median profile across images.
Full protocol, model and probe-set specifications, and per-model tables are given in the appendix \appendixref{appendix:A}; per-image profiles across all evaluation sets are shown in \cref{app:profiles}. 

\begin{figure}[t]
    \includegraphics[width=0.94\linewidth]{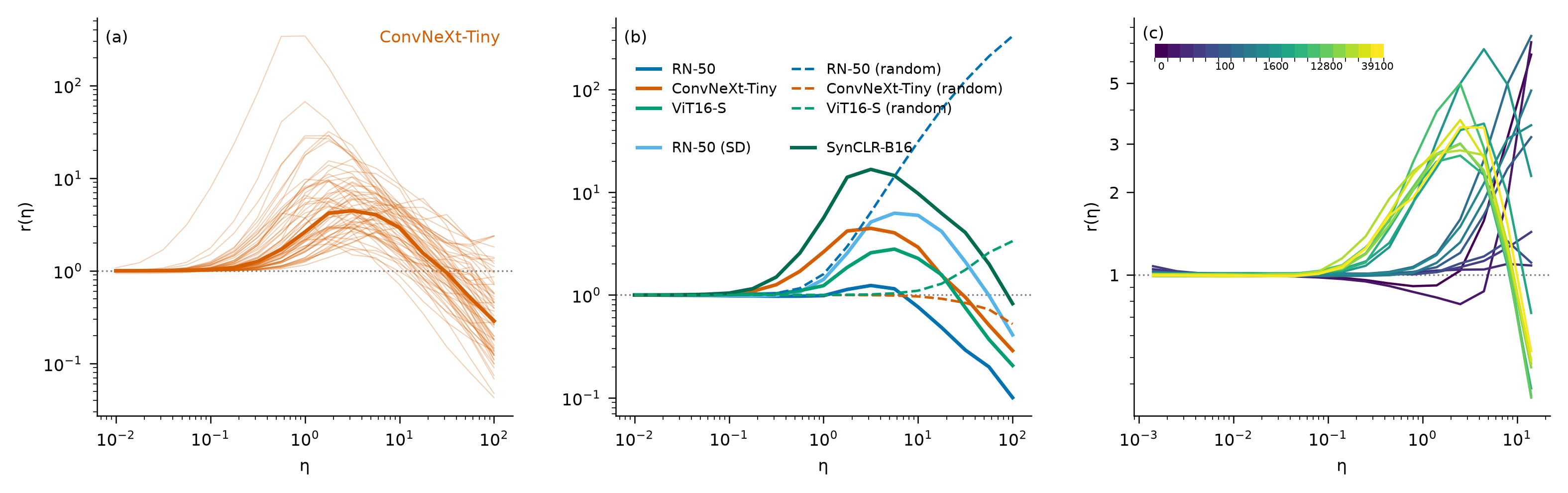}
    \caption{
        Scale-resolved response profiles. The horizontal axis is the perturbation
    magnitude $\eta$, and the vertical axis is $r_\eta$, the ratio of the
    measured feature displacement to its local linear prediction.
    (a) Individual profiles of single images are thin and their median is bold (ConvNeXt-Tiny encoder).
    (b) Across diverse encoder architectures, trained models (solid) exhibit
    the characteristic plateau--rise--peak--decay \emph{bump}, whereas
    randomly initialized models (dashed) do not (medians reported).
    (c) During training, the bump forms early and remains largely
    stable.
    }
    
    \label{fig:main}
\end{figure}

\paragraph{A characteristic bump after training.}
We first evaluate frozen encoders spanning convolutional and transformer
architectures, standard and adversarial ImageNet supervision,
synthetic-data training, and random initialization. At the level of
individual images, the profiles vary in magnitude but retain a consistent
overall shape (\cref{fig:main}a). Across all trained encoders, the median
profiles exhibit an initial plateau near $r_\eta=1$, followed by a rise
above the local prediction, an interior peak, and an eventual decay
(\cref{fig:main}b). We call this structure the \emph{bump}. No randomly
initialized encoder exhibits a comparable peak, suggesting that training is a required (but not sufficient) condition for the phenomenon. The bump's height, location, and width nevertheless vary across models and evaluation data.

\paragraph{Formation during training.}
To observe how the bump emerges, we train ResNet-18, VGG-16, and WRN-28-10 classifiers on CIFAR-10 and measure their profiles throughout training.
In every architecture, the bump is absent at initialization, emerges early,
and is subsequently largely preserved (\cref{fig:main}c). It appears for
both training and validation images, indicating that it is not specific to
memorized examples.

\begin{figure}
    \includegraphics[width=0.94\linewidth]{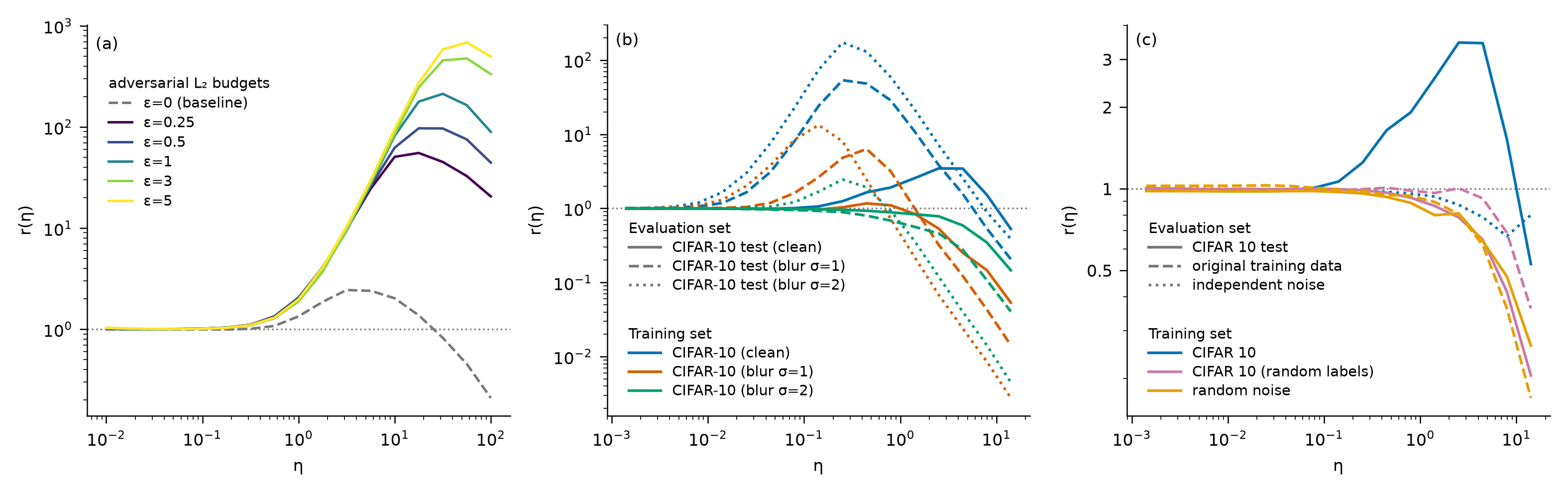}
    \caption{
    Response profiles under controlled training conditions.
    (a) Adversarial training moves the peak $10$--$30\times$ outward, primarily through an approximately $2000\times$ reduction in the local prediction.
    (b) ResNet-18 trained on clean or blurred ($\sigma \in \{1, 2\}$) CIFAR-10 images evaluated on clean or blurred test images. 
    The bump mostly grows with evaluation blur, and shrinks with training blur.
    (c) Profiles for ResNet-18 trained on CIFAR-10 with randomized labels and random data show no peak on any evaluation set. 
    }
    \label{fig:ablation}
\end{figure}

\paragraph{Controlled training conditions.}
We next examine how the profile changes under modified training settings.
Adversarial training moves the peak to substantially larger perturbation
scales, primarily through a reduction in the locally predicted response
(\cref{fig:ablation}a). Training and evaluating on blurred CIFAR-10 images
systematically changes the bump, demonstrating its dependence on the
relation between the training and evaluation distributions
(\cref{fig:ablation}b). Finally, under randomized labels, the model reaches $89.5\%$ training
accuracy while remaining at chance-level test accuracy ($10.6\%$),
indicating substantial memorization without generalization. Nevertheless,
no comparable peak forms under either randomized labels or random-noise
inputs (\cref{fig:ablation}c), showing that substantial memorization alone
is insufficient to produce the bump. Together, these
results associate the bump with meaningful input--target structure and
show that its shape reflects both the training distribution and objective.

\section{Conclusions and Limitations}
\label{sec:04}

We introduced a scale-resolved measure of how an encoder's finite response
departs from its local linear geometry. Across diverse image encoders, we
identified the \emph{bump}, a geometric structure that emerges during
meaningful training and is empirically associated with learning shared structure that transfers to unseen examples. 

Our measure captures only response magnitude averaged over isotropic
directions; at larger scales, these perturbations might leave the natural-image manifold.
Moreover, the observed connection to learning and generalization is
empirical and does not establish the bump's causal mechanism. Extending the
analysis beyond image classification remains future work.

\bibliography{refs.bib}

\appendix
\section{}
\label{appendix:A}

\subsection{Theoretical details}
\label{app:theory}

In this section, we provide a more detailed overview of the method introduced in \cref{sec:02}.

\paragraph{Setup and the pullback metric.}
Consider the space $\mathcal{I} \simeq \R^{d_{\mathcal{I}}}$, $d_{\mathcal{I}} = 3 H W$, of flattened $H \times W$ RGB image tensors.
An image encoder is a map $\Phi : \mathcal{I} \to \mathcal{Z}$ taking an image $\bx$ to its $d_{\mathcal{Z}}$-dimensional feature representation $\Phi(\bx) \in \mathcal{Z} \simeq \R^{d_{\mathcal{Z}}}$.
Both spaces $\mathcal{I}, \mathcal{Z}$ are smooth manifolds.
Following the standard assumption that two feature vectors $\bz, \bz' \in \mathcal{Z}$ can be compared using the inner product $\bz^\top \bz'$, we equip $\mathcal{Z}$ with a flat Riemannian metric structure – that is, we equip each tangent space $\T_{\bz} \mathcal{Z}$ with the inner product $\bm{\xi}_1^\top \bm{\xi}_2$ for $\bm{\xi}_1, \bm{\xi}_2 \in \T_{\bz} \mathcal{Z}$.
When $\Phi$ is differentiable at $\bx \in \mathcal{I}$, its differential at that point, $\mathrm{D}_{\bx} \Phi : \T_{\bx}\mathcal{I} \to \T_{\Phi(\bx)}\mathcal{Z}$, is represented in coordinates by the Jacobian $\J := \J_\Phi(\bx) \in \R^{d_{\mathcal{Z}} \times d_{\mathcal{I}}}$.
This map takes an image-space direction $\bu \in \T_{\bx}\mathcal{I}$ to the feature-space direction $\J \bu$ along which the representation moves as $\bx$ moves along $\bu$.

To compare image-space tangent vectors at $\bx$ by the effect they induce on the features $\Phi(\bx)$, we pull back the feature-space flat metric along $\Phi$.
Concretely, this results in a bilinear form $g_{\bx} : \T_{\bx} \mathcal{I} \times \T_{\bx} \mathcal{I} \to \R$ on each tangent space 
$$
    g_{\bx}(\bu_1, \bu_2)
    = [\J \bu_1]^{\top} [\J \bu_2]
    = \bu_1^{\top} [\J^{\top} \J]\, \bu_2 ,
$$
defined wherever $\Phi$ is differentiable.
The assignment $\bx \mapsto g_{\bx}$ is known as the pullback metric \citep{lee2018riemannian}, and the matrix $\G := \G_{\Phi}(\bx) = \J^{\top} \J$ is its coordinate representation.
Since $d_{\mathcal{Z}} < d_{\mathcal{I}}$ for essentially all encoders, the pullback metric is degenerate: $\G$ is positive semi-definite with $\bu_2^{\top} \G \bu_1 = 0$ whenever $\bu_1 \in \ker \J$ or $\bu_2 \in \ker \J$, and $\dim \ker \J \ge d_{\mathcal{I}} - d_{\mathcal{Z}}$.

\paragraph{The ratio.}
The pullback metric describes the encoder's local geometry at an input $\bx$.
As introduced in \cref{sec:02}, we measure how far this description remains predictive at finite perturbation scales.
For a unit tangent vector $\bu \in \mathbb{S}^{d_{\mathcal{I}} -1} \subset \T_{\bx} \mathcal{I}$, let 
$$
    D_{\eta}(\bx, \bu) := \| \Phi(\bx + \eta \bu) - \Phi(\bx) \|_2^2 
$$
denote the squared displacement in the features induced by perturbing $\bx \to \bx + \eta\bu$ at a scale $\eta > 0$.
The pullback metric predicts this displacement as the squared length of the first-order feature displacement $\mathrm{D}_{\bx}\Phi\,(\eta\bu) = \eta\J\bu \in \T_{\Phi(\bx)} \mathcal{Z}$, that is
$$
    \tilde{D}_{\eta}(\bx, \bu) 
    := (\eta \bu)^{\top} \G (\eta \bu)
    = \eta^2 \| \J \bu \|_2^2.
$$

The actual displacement and its prediction need not agree.
Moreover, comparing $D_{\eta}(\bx, \bu)$ and $\tilde{D}_{\eta}(\bx, \bu)$ requires choosing a direction along which we probe the representation, whereas we seek a description of its behavior at $\bx$ that depends solely on the encoder at the point at which we measure.
We opt to remove this directionality by simply averaging both quantities over the uniform measure on the sphere.
Let 
$$
    m_\eta(\bx) := \E_{\bu \sim \operatorname{Unif}(\mathbb{S}^{d_{\mathcal{I}}-1})} \left[ D_\eta(\bx, \bu) \right],
    \quad
    \tilde{m}_\eta(\bx) := \E_{\bu \sim \operatorname{Unif}(\mathbb{S}^{d_{\mathcal{I}}-1})} \big[ \tilde{D}_\eta(\bx, \bu) \big] = \eta^2 \operatorname{tr} \G \, / \, d_{\mathcal{I}}
$$
denote the mean response and the mean prediction at scale $\eta$, respectively.
We define their ratio as
$$
    r_{\eta}(\bx) = \frac{m_\eta(\bx)}{\tilde{m}_{\eta}(\bx)},
$$
recovering \cref{eq:r_eta}.

\paragraph{Taylor expansion and sampling.}
Suppose $\Phi$ is smooth in a neighborhood of $\bx$.
Expanding to second order,
$$
    \Phi(\bx + \eta\bu) - \Phi(\bx)
    = \eta \J \bu + \frac{\eta^2}{2} \mathbf{H}[\bu, \bu] + O(\eta^3),
$$
where $\mathbf{H}$ is the second differential of $\Phi$ at $\bx$, a symmetric bilinear map taking values in the feature tangent space $\T_{\Phi(\bx)}\mathcal{Z}$.
Taking the squared norm yields
$$
    D_{\eta}(\bx, \bu)
    = \tilde{D}_{\eta}(\bx, \bu) + \eta^3 [\J \bu]^{\top} \mathbf{H}[\bu,\bu] + O(\eta^4).
$$
The map $\bu \mapsto \J \bu$ is linear (odd), and $\mathbf{H}$ is a quadratic (even) function of $\bu$, so $\bu \mapsto [\J \bu]^{\top} \mathbf{H}[\bu,\bu]$ is odd.
Since the uniform sphere measure is invariant under rotations, 
this term vanishes when we take the expectation, leaving us with:
$$
    m_{\eta}(\bx) = \tilde{m}_{\eta}(\bx) + O(\eta^4),
$$
and consequently $r_{\eta}(\bx) = 1 + O(\eta^2)$ – the ratio departs quadratically from unity with no linear term.
This's crucial for our measurements: a linear term would bias the reading of the departure from $r_{\eta} = 1$ toward smaller $\eta$.

However, in practice, both means are estimated from a finite set of directions $U \subset \mathbb{S}^{d_{\mathcal{I}}-1}$.
For a sample of independently drawn directions, the empirical average of the odd term
$$
    \frac{1}{|U|} \sum_{\bu \in U} [\J \bu]^{\top} \mathbf{H}[\bu, \bu]
$$
is a mean of $|U|$ independent zero-mean random variables.
It vanishes only in expectation, and its typical size for any single draw is of order $|U|^{-1/2}$.
The estimated ratio thus retains a term linear in $\eta$ — precisely the bias described above.
To address this issue, we sample each direction $\bu \in U$ together with its antipode $-\bu$, making $U$ closed under negation: $U = -U$.
The sum of any odd function over such $U$ cancels pairwise, and is exactly zero for every draw rather than merely in expectation.
The estimate then departs from its small-$\eta$ level at order $\eta^2$, matching the earlier derivation.

Note that for piecewise-affine encoders such as ReLU networks, no expansion is needed: within the activation region containing $\bx$ the encoder is exactly affine, so $r_{\eta}(\bx) = 1$ holds exactly until the perturbation reaches outside the region boundary.

\subsection{Experimental details}
\label{app:details}

This section provides further detail on the experimental protocol used in \cref{sec:03}.

\paragraph{Encoders.} 
We evaluate nine ImageNet encoders: ResNet-50 \citep{he2015deepresiduallearningimage} with torchvision \texttt{IMAGENET1K\_V2} weights ($d_\mathcal{Z} = 2048$), ConvNeXt-Tiny \citep{liu2022convnet2020s} with torchvision weights ($d_\mathcal{Z} = 768$), ViT-S/16 \citep{dosovitskiy2021imageworth16x16words} with timm \texttt{augreg\_in1k} weights ($d_\mathcal{Z} = 384$), the ImageNet-SD ResNet-50 of \citet{sariyildiz2023faketillmakeit} ($d_\mathcal{Z} = 2048$), the adversarially trained $L_2$ ResNet-50 checkpoints of \citet{salman2020adversariallyrobustimagenetmodels} ($d_\mathcal{Z} = 2048$), SynCLR-B/16 \citep{tian2023learningvisionmodelsrivals}, a ViT-B/16 trained on synthetic images ($d_\mathcal{Z} = 768$), as well as randomly initialized copies of ResNet-50, ConvNeXt-Tiny, and ViT-S/16.
We additionally train ResNet-18 \citep{he2015deepresiduallearningimage},
VGG-16 \citep{simonyan2015deepconvolutionalnetworkslargescale}, and
WRN-28-10 \citep{zagoruyko2017wideresidualnetworks} classifiers on
CIFAR-10 \citep{cifar} and measure them under the same protocol with a
perturbation ladder adjusted to the input resolution; training
configurations are given below.

\paragraph{Evaluation sets.}
Each evaluation set is a fixed collection of images.
For ImageNet-resolution measurements, \texttt{val64} contains one image per each of 64 seeded classes from the ImageNet-1k validation split \citep{russakovsky2015imagenetlargescalevisual}, taking the lexicographically first validation image of each class; \texttt{a64} and \texttt{o64} are seeded 64-image draws from ImageNet-A and ImageNet-O \citep{Hendrycks_2021_CVPR}; \texttt{sd64} contains one Stable Diffusion~1.4 \citep{rombach2022highresolutionimagesynthesislatent} render per \texttt{val64} class; and \texttt{blur64} applies Gaussian blur to \texttt{val64} ($\sigma = 2$ at 512\,px before the resize to measurement resolution, $\sigma \approx 1$ in the measured frame).
For CIFAR-10, \texttt{cifar32} and \texttt{cifar32train} are seeded 32-image draws from the test and train splits, \texttt{cifar32b1} and \texttt{cifar32b2} blur the \texttt{cifar32} images ($\sigma = 1, 2$), \texttt{noise32} contains 32 i.i.d.\ uniform noise images, and \texttt{noise32train} regenerates 32 training inputs of the random-data network from its data seed.
Main-text figures report \texttt{val64} unless stated otherwise; \appendixref{app:profiles} shows profiles across all sets.

\paragraph{Protocol.}
Each profile is measured on a ladder of 17 logarithmically spaced perturbation scales, spanning $10^{-2}$–$10^{2}$ at ImageNet resolution and $10^{-2.85}$–$10^{1.15}$ on CIFAR-10, the latter shifted to match the former in per-pixel amplitude.
At each scale, the response is averaged over 2048 unit directions per image, drawn as 1024 Gaussian vectors normalized to unit norm and extended with their antipodes, from a fixed per-image seed shared across all scales.
Because the linear prediction concentrates quickly, it is computed from only 256 exact Jacobian--vector products on the first directions of the same set.
Perturbations are applied in raw $[0,1]$ pixel space before each model's input normalization, which is folded into $\Phi$, so $\eta$ carries the same meaning for every encoder.
Perturbed inputs are not projected back to $[0,1]$, so at large $\eta$ they leave the valid pixel range.
For every image, $r_\eta$ is evaluated per scale, and the reported profiles are medians across the evaluation set.
At the smallest scales, the measured displacement can approach floating-point resolution.
Thus rungs where estimated floating-point noise exceeds 5\% of the measured energy are discarded per image, and the median at each scale is taken over the remaining images.

\paragraph{Training.}
The CIFAR-10 classifiers are trained with SGD (momentum $0.9$, weight decay $5 \times 10^{-4}$), batch size $128$, initial learning rate $0.1$ with per-step cosine decay to zero, for $100$ epochs ($39100$ steps), with random crop and horizontal flip augmentation.
For the randomized-label variant, the training labels are randomly permuted across the entire training set, destroying the image--label pairing while preserving the label counts; for the random-data variant, the training set is replaced by seeded uniform noise images; both variants disable augmentation.
The blur-trained variants apply Gaussian blur ($\sigma \in \{1, 2\}$) to the training images once, before the standard pipeline.
For the formation experiment, checkpoints are saved on an approximately logarithmic step schedule from initialization to the end of training and measured under the protocol above.
All measurements were run on a single RTX 4080; all random draws in training, data construction, and measurement are seeded.

\subsection{Related works}
\label{app:related}

In this section, we position our work in the literature.

\paragraph{Applications of the pullback metric.}
The pullback metric has emerged as a standard tool for studying the latent space of deep generative models. When the decoder is an immersion, the pullback of the ambient Euclidean metric yields a Riemannian metric on the model's latent space.
Shortest paths under this metric yield distances and interpolants that follow the data manifold \citep{tosi2014metricsprobabilisticgeometries, arvanitidis2021latent, chen2018metricsdeepgenerativemodels, shao2017riemanniangeometrydeepgenerative}, and its volume element serves as a distortion diagnostic and regularizer \citep{nazari2023geometricautoencodersdecode}.
In perceptual modeling, pulliback back the Fisher-Rao metric of a stochastic response model instead yields the Fisher information on stimulus space, whose extremal eigenvectors predict the most- and least-noticeable image distortions \citep{berardino2018eigendistortionshierarchicalrepresentations, feather2025discriminatingimagerepresentationsprincipal}.

\paragraph{Jacobian-based sensitivity statistics.}
In motivating their sensitivity measure, \citet{novak2018sensitivity} approximate the expected squared change of the network output under a small isotropic Gaussian perturbation of the input by its first-order expansion.
In our notation:
$$
\E_{\Delta\bx}\left[\|\Phi(\bx+\Delta\bx)-\Phi(\bx)\|^{2}\right]
\approx
\E_{\Delta\bx}\left[\|\J_{\Phi}(\bx)\Delta\bx\|^{2}\right]
=
\varepsilon\|\J_{\Phi}(\bx)\|_{F}^{2}
$$
for $\Delta\bx\sim\mathcal{N}(0,\varepsilon I)$.
With the Gaussian measure replaced by the uniform sphere measure, writing $\Delta \bx = \eta \bu$, the left hand side is the numerator of $r_{\eta}$, and the right-hand side is its denominator.
They use this approximation as a premise for infinitesimal
perturbations; only the right-hand side is kept, as a pointwise
statistic, and where it holds is never evaluated.
Our approach takes this approximate equality as its starting point
and evaluates its validity at finite $\eta$, across scales.

\paragraph{Finite-scale linearity and linear regions.}
\citet{qin2019adversarialrobustnesslocallinearization} penalize the largest violation of the first-order Taylor expansion of the loss over a perturbation ball, as a regularizer for adversarial training.
Their quantity is built from the scalar loss rather than the representation map, and it shapes training, while our $r_{\eta}$ constitutes only a measurement.
Moreover, a maximum over a ball is a nondecreasing function of its radius by construction, so its profile cannot highlight a peak.
\citet{humayun2024deepnetworksgrok} count the linear-region boundaries of ReLU networks in a small fixed-radius neighborhood of a point and track the count over training, observing a descent, an ascent, and a second descent in which boundaries migrate away from the data.
Region counting of this kind builds on a line of work bounding the
number and local density of linear regions in deep networks
\citep{hanin2019complexitylinearregionsdeep, hanin2019deeprelunetworkssurprisingly}.

\paragraph{Random networks and the NTK.}
The monotone rise without turnover that we observe at random initialization (Figure~\ref{fig:main}) is consistent with the expansivity of random deep networks, whose input trajectories lengthen exponentially with depth \citep{poole2016exponentialexpressivitydeepneural, raghu2017expressivepowerdeepneural}.
The linearization studied in the neural tangent kernel literature is in
the parameters, over training \citep{jacot2020neuraltangentkernelconvergence, chizat2020lazytrainingdifferentiableprogramming}, whereas the one measured in this work is in the input, for a network whose parameters are frozen during the measurement.
\clearpage
\section{Full response profiles}
\label{app:profiles}

\begin{figure}[h]
\centering
\includegraphics[width=\textwidth]{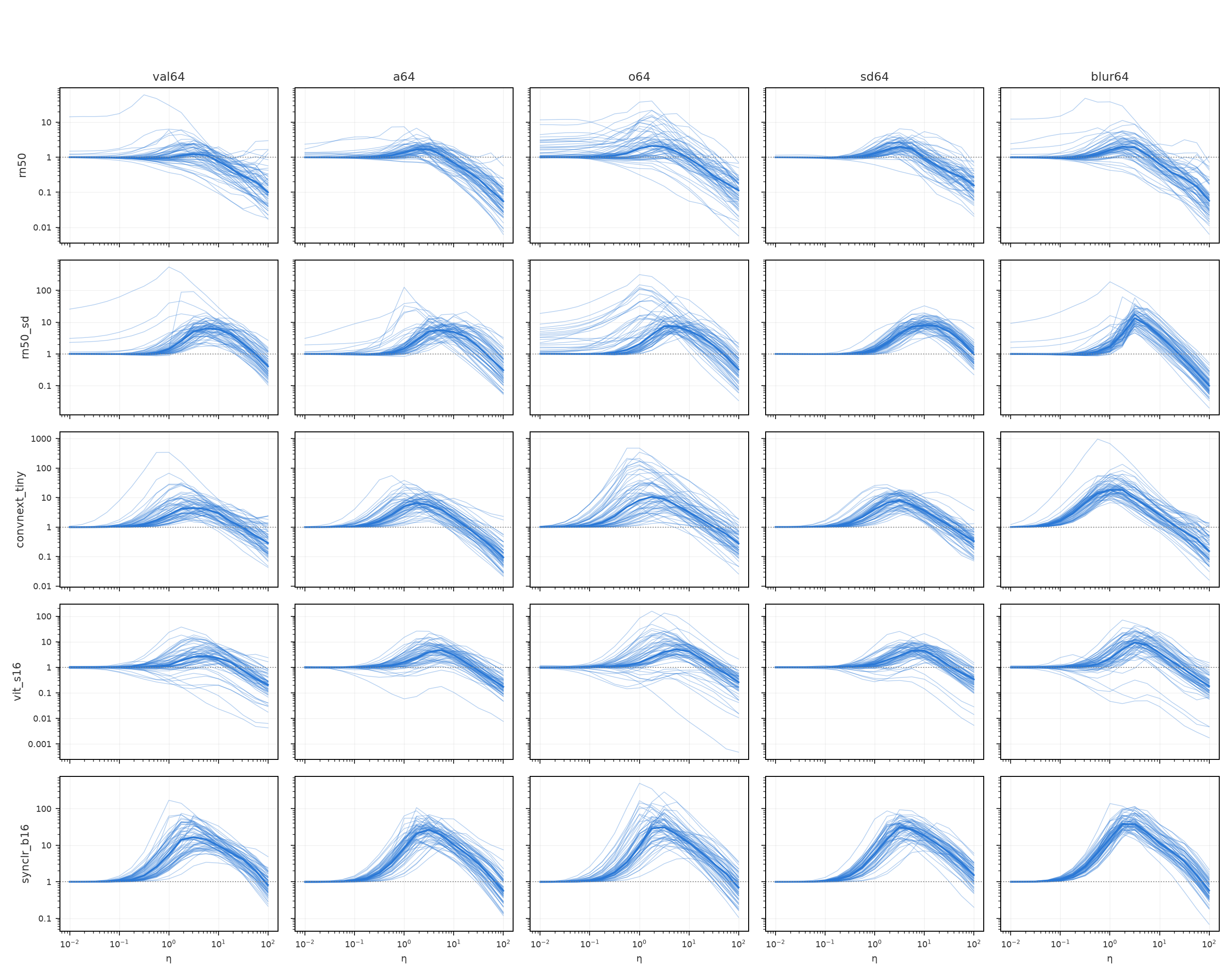}
\caption{Per-image response profiles for the trained ImageNet encoders
(rows) on the five ImageNet-resolution evaluation sets (columns). Each
line is one image. The plateau--rise--peak--decay shape persists across
encoders and evaluation sets; height and location vary.}
\label{fig:zoo-sets}
\end{figure}

\begin{figure}[h]
\centering
\includegraphics[width=\textwidth]{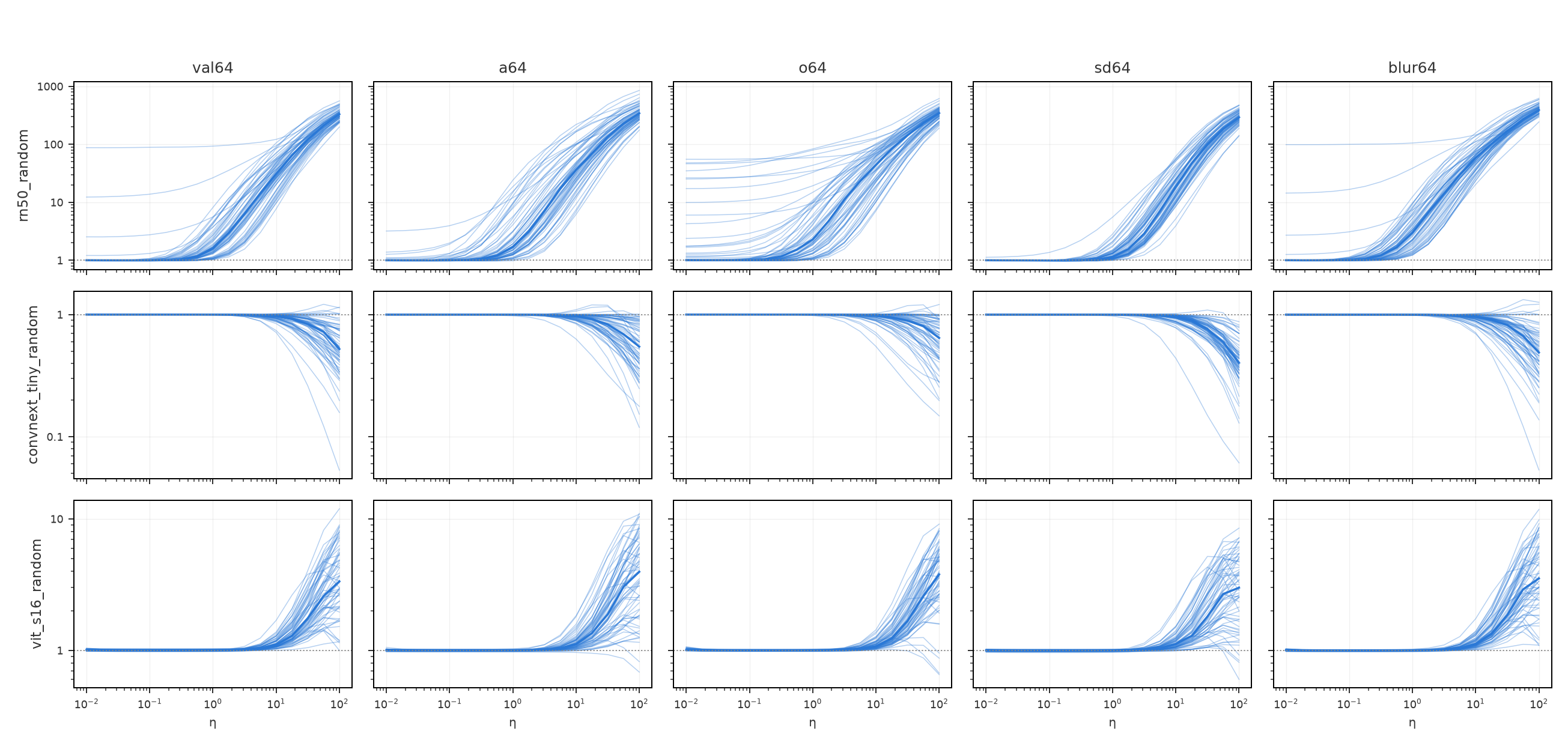}
\caption{Per-image response profiles for the randomly initialized
controls. No interior peak forms on any evaluation set: RN-50 and
ViT-S/16 rise monotonically to the end of the ladder, and ConvNeXt-Tiny
stays near $r_\eta = 1$ before decaying}
\label{fig:rand-sets}
\end{figure}

\begin{figure}[h]
\centering
\includegraphics[width=\textwidth]{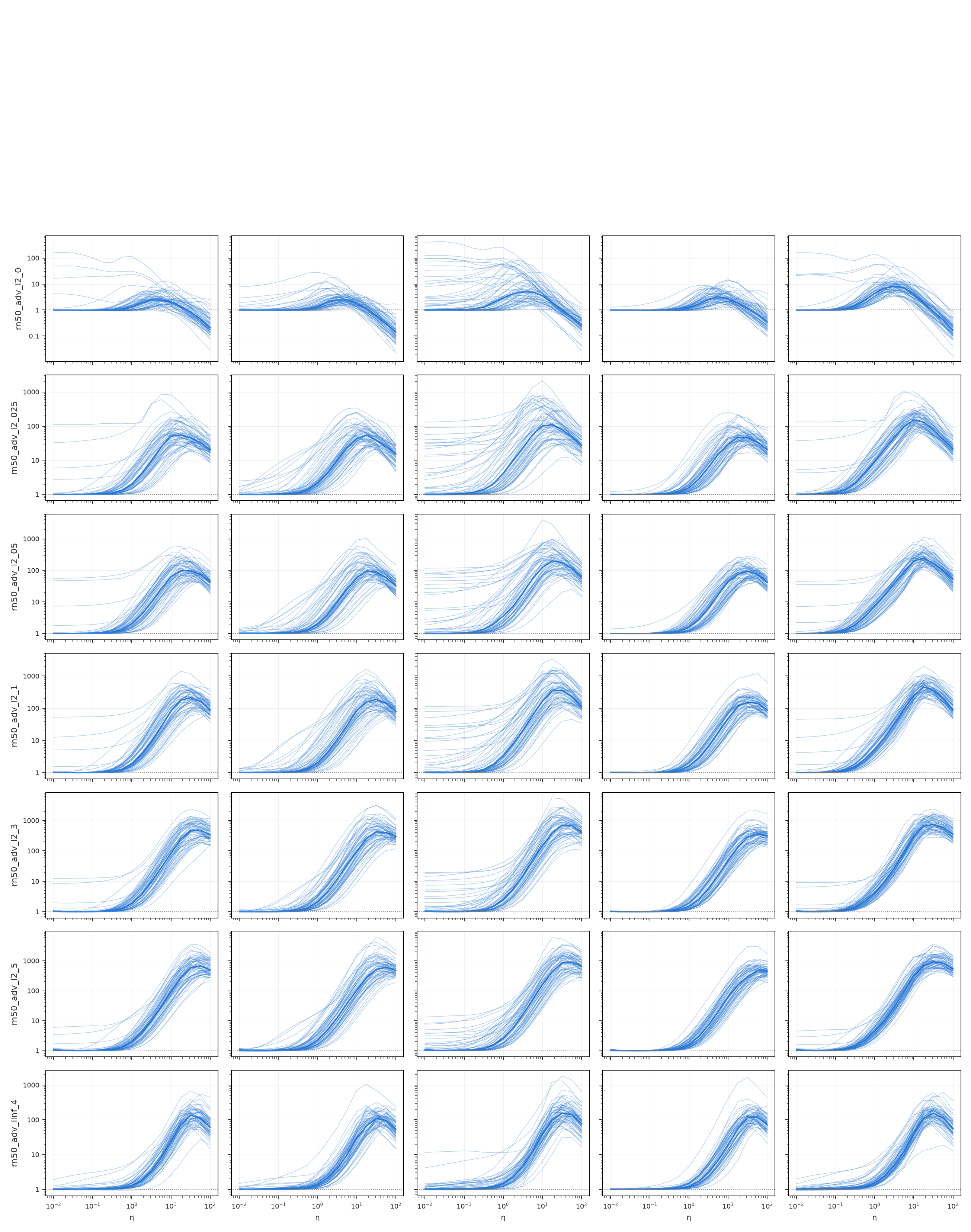}
\caption{Per-image response profiles for the adversarially trained
ResNet-50 family. With increasing training budget $\varepsilon$, the
peak moves to larger $\eta$ and grows; the baseline
$\varepsilon = 0$ baseline retains a peak at small scale.}
\label{fig:adv-sets}\label{jmlrend}
\end{figure}

Figures~\ref{fig:zoo-sets}, \ref{fig:rand-sets} and~\ref{fig:adv-sets} 
show per-image profiles for the trained encoders, the
randomly initialized controls, and the adversarially trained family,
across all five ImageNet-resolution evaluation sets.

\end{document}